\pdfoutput=1

\documentclass[11pt]{article}

\usepackage[final]{acl}

\usepackage{times}
\usepackage{latexsym}

\usepackage[T1]{fontenc}

\usepackage[utf8]{inputenc}

\usepackage{microtype}

\usepackage{inconsolata}

\usepackage{graphicx}
\usepackage{tcolorbox}
\tcbuselibrary{breakable}

\usepackage{multirow}

\usepackage{amsmath, amssymb}
\DeclareMathOperator{\NAR}{NAR}
\DeclareMathOperator{\HAR}{HAR}

\title{Is External Information Useful for Stance Detection with LLMs?}

\author{Quang Minh Nguyen \\
  KAIST \\
  Daejeon, South Korea \\
  \texttt{qm.nguyen@kaist.ac.kr} \\\And
  Taegyoon Kim \\
  KAIST \\
  Daejeon, South Korea \\
  \texttt{taegyoon@kaist.ac.kr} \\
  }

\begin{document}
\maketitle
\begin{abstract}
In the stance detection task, a text is classified as either favorable, opposing, or neutral towards a target. Prior work suggests that the use of external information, e.g., excerpts from Wikipedia, improves stance detection performance. However, whether or not such information can benefit large language models (LLMs) remains an unanswered question, despite their wide adoption in many reasoning tasks. In this study, we conduct a systematic evaluation on how Wikipedia and web search external information can affect stance detection across eight LLMs and in three datasets with 12 targets. Surprisingly, we find that such information degrades performance in most cases, with macro F1 scores dropping by up to 27.9\%. We explain this through experiments showing LLMs' tendency to align their predictions with the stance and sentiment of the provided information rather than the ground truth stance of the given text. We also find that performance degradation persists with chain-of-thought prompting, while fine-tuning mitigates but does not fully eliminate it. Our findings, in contrast to previous literature on BERT-based systems which suggests that external information enhances performance, highlight the risks of information biases in LLM-based stance classifiers\footnote{Code is available at \href{https://github.com/ngqm/acl2025-stance-detection}{https://github.com/ngqm/acl2025-stance-detection}}.
\end{abstract}

\section{Introduction}

Stance detection is a task that determines whether a given content supports, opposes, or remains neutral toward a target. When the content assumes implicit information about the target, stance detection systems can benefit from external information, such as Wikipedia excerpts, regarding the target. Accordingly, recent research has explored incorporating such information to improve stance detection, highlighting its benefits \citep{bartstance, backgroundstance, infusing, enhancing}.

On the other hand, large language models (LLMs) have demonstrated remarkable capabilities across various reasoning tasks, including mathematical reasoning~\citep{mathprompter}, coding~\citep{deepseek-coder}, and language understanding~\citep{emergent}. Given these advances, recent research has begun exploring the potential of LLMs for stance detection~\citep{tree-of-counterfactual, role-infused}.

With these parallel trends, an important question arises: \textit{Can external information enhance LLMs in stance detection?} In this paper, we systematically evaluate how external information about targets impacts the performance of a diverse set of LLMs across a wide range of datasets and targets. 

Surprisingly, we find that such information, from Wikipedia or web search, tends to \textit{compromise} stance detection performance.
To explain this phenomenon, we show that model predictions often adopt external information stance and sentiment. 
Despite LLMs' known sensitivity to prompt variations, we show consistent results through experiments on different chain-of-thought prompting methods. Finally, we find that fine-tuning mitigates but does not fully eliminate performance degradation. Our research serves as a caution against the use of external information without proper bias consideration for LLMs in stance detection and natural language reasoning at large. 

\section{Related Work}

\textbf{Stance Detection with External Information.}~
A key line of related work investigates leveraging external information, often from Wikipedia, to enhance stance detection. \citet{infusing} fine-tuned BERT models which take Wikipedia excerpts, in addition to given texts and targets, as inputs and report significantly improved stance detection performance. Subsequent works in the literature either utilized external information in a different formulation of stance detection~\citep{bartstance} or introduced new 
knowledge organization and filtering schemes for such information~\citep{backgroundstance, enhancing}.
While these works have primarily focused on fine-tuning smaller, BERT-like models for stance detection, we extend this research to LLMs, which possess emergent reasoning abilities but require significantly more resources for fine-tuning.

\noindent \textbf{Stance Detection with LLMs.}~
Relatedly, another stream of works examines how LLMs can be applied to stance detection. \citet{tree-of-counterfactual} and \citet{role-infused} proposed prompting schemes where reasoning on stance is organized as ensembles or multi-agent discussions. Meanwhile, \citet{calibration} introduced a calibration network which serves to mitigate internal biases of LLMs. 
Orthogonal yet complementary to these efforts, our work provides a foundational analysis of how external information influences their decision-making, uncovering unintended effects and offering insights to guide future research.

\section{Experimental Setup}

\subsection{Data, Model, and Metric}

We utilize the following datasets, which are all in English and widely used in stance detection research.
\begin{enumerate}
    \item COVID-19-Stance \cite{covid19stance}: 6,133 Tweets about COVID-19 in the U.S.: Fauci, school closure, stay-at-home orders, and face masking. Labels are either FAVOR, AGAINST, or NONE.
    \item P-Stance \cite{pstance}: 21,574 Tweets with Trump, Biden, and Sanders as targets. Labels are either FAVOR or AGAINST.
    \item SemEval 2016 Task 6 \cite{semeval}: 4,163 Tweets about atheism, climate change, feminist movement, Hillary Clinton, and abortion. Labels are either FAVOR, AGAINST, or NONE.
\end{enumerate}

For our experiments, we consider a total of eight popular LLMs of various sizes, both open- and closed-source (see Table \ref{tab:accuracy-f1-change}). We utilize the instruction-tuned versions of all open-source models (the "-Instruct" postfix omitted in the paper). Additionally, we use WS-BERT \citep{infusing} as a BERT baseline. Since stance detection is a task requiring determinism over creativity, we set the inference temperature of all models to zero; as a result, each run produced outputs with negligible variation in performance, and our results are therefore based on single runs.
We evaluate models through accuracy and macro F1. More details on data, models, prompts, and output validation are in Appendices~\ref{app:llms}, \ref{app:prompts}, and \ref{app:validation}. 

\subsection{External Information}

Following previous work, we utilize external information from Wikipedia collected by \citet{infusing} for COVID-19-Stance and P-Stance. The external information for SemEval 2016 Task 6 was collected ourselves through the Wikipedia API. Additionally, we consider both {\it synthetic} and {\it real-world} sources of bias: the former is implemented through Against and Favor stance injection with GPT-4o mini, while the latter employs information fetched through OpenAI's Web Search API with GPT-4o mini~\citep{web-search}, which may reflect non-objective content retrieved from the web. The stance injection and web search prompts are included in Appendix~\ref{app:prompts}.

\begin{table*}[!h]
    \centering
    \begin{tabular}{cccccc}
	\hline 
	Model & No Info & Wiki & Web & Wiki (Against) & Wiki (Favor)\\ 
	\hline 
	\multicolumn{6}{c}{COVID-19-Stance (Accuracy~/~Macro F1 in \%)}\\ 
 	\hline 
	Llama-3.2-3B & 49.4 / 44.8 & {\color{red} -12.4} / {\color{red} -14.9} & {\color{red} -12.9} / {\color{red} -14.8} & {\color{red} -17.4} / {\color{red} -26.2} & {\color{red} -9.5} / {\color{red} -15.8}\\ 
	Llama-3.1-8B & 64.1 / 63.5 & {\color{red} -0.2} / {\color{red} -1.2} & {\color{red} -6.2} / {\color{red} -8.3} & {\color{red} -4.1} / {\color{red} -5.8} & {\color{red} -1.9} / {\color{red} -3.1}\\ 
	Qwen2.5-3B & 59.6 / 54.3 & {\color{red} -6.4} / {\color{red} -2.6} & {\color{red} -15.6} / {\color{red} -14.3} & {\color{red} -10.2} / {\color{red} -9.0} & {\color{red} -10.6} / {\color{red} -10.3}\\ 
	Qwen2.5-7B & 63.5 / 58.1 & {\color{red} -3.4} / {\color{red} -6.1} & {\color{red} -4.9} / {\color{red} -5.7} & {\color{red} -4.8} / {\color{red} -8.0} & {\color{red} -2.4} / {\color{red} -6.0}\\ 
	Qwen2.5-14B & 72.5 / 70.8 & {\color{red} -3.2} / {\color{red} -3.1} & {\color{red} -3.6} / {\color{red} -3.8} & {\color{red} -5.9} / {\color{red} -11.1} & {\color{red} -1.9} / {\color{red} -6.8}\\ 
	Qwen2.5-32B & 72.0 / 71.0 & {\color{red} -3.2} / {\color{red} -3.8} & {\color{red} -2.6} / {\color{red} -3.1} & {\color{red} -6.1} / {\color{red} -7.2} & {\color{red} -1.1} / {\color{red} -1.7}\\ 
	GPT-4o mini & 64.1 / 62.9 & {\color{red} -9.1} / {\color{red} -9.4} & {\color{red} -10.9} / {\color{red} -11.8} & {\color{red} -12.0} / {\color{red} -14.9} & {\color{red} -11.9} / {\color{red} -12.3}\\ 
	Claude 3 Haiku & 64.4 / 63.7 & {\color{red} -7.5} / {\color{red} -9.0} & {\color{red} -11.4} / {\color{red} -18.3} & {\color{red} -14.0} / {\color{red} -16.9} & {\color{red} -7.9} / {\color{red} -12.4}\\ 
	WS-BERT & 83.9 / 82.7 & {\color{blue} +0.2} / {\color{blue} +0.2} & {\color{blue} +0.1} / {\color{blue} +0.1} & {\color{blue} +0.4} / {\color{blue} +0.4} & {\color{blue} +0.2} / {\color{blue} +0.2}\\ 
	\hline 
	\multicolumn{6}{c}{P-Stance (Accuracy~/~Macro F1 in \%)}\\ 
 	\hline 
	Llama-3.2-3B & 77.4 / 59.0 & {\color{red} -12.6} / {\color{red} -15.7} & {\color{red} -18.4} / {\color{red} -16.5} & {\color{red} -19.7} / {\color{red} -21.5} & {\color{red} -20.9} / {\color{red} -22.5}\\ 
	Llama-3.1-8B & 81.7 / 63.0 & {\color{red} -5.3} / {\color{blue} +12.1} & {\color{red} -6.0} / {\color{blue} +11.5} & {\color{red} -13.4} / {\color{red} -0.1} & {\color{red} -9.8} / {\color{blue} +7.8}\\ 
	Qwen2.5-3B & 65.4 / 64.4 & {\color{red} -9.2} / {\color{red} -14.0} & {\color{red} -6.4} / {\color{red} -14.0} & {\color{red} -1.7} / {\color{red} -19.1} & {\color{red} -3.8} / {\color{red} -6.7}\\ 
	Qwen2.5-7B & 74.3 / 48.4 & {\color{red} -1.4} / {\color{red} -1.6} & {\color{red} -0.2} / {\color{blue} +6.0} & {\color{red} -8.0} / {\color{red} -8.0} & {\color{red} -0.1} / {\color{red} -0.3}\\ 
	Qwen2.5-14B & 81.7 / 54.2 & {\color{red} -1.8} / {\color{blue} +7.1} & {\color{red} -1.3} / {\color{blue} +7.3} & {\color{red} -2.8} / {\color{blue} +6.1} & {\color{red} -1.0} / {\color{blue} +8.0}\\ 
	Qwen2.5-32B & 84.3 / 65.2 & {\color{red} -0.8} / {\color{blue} +8.2} & {\color{red} -0.6} / {\color{blue} +8.4} & {\color{red} -2.4} / {\color{blue} +15.9} & {\color{red} -0.3} / {\color{blue} +8.8}\\ 
	GPT-4o mini & 80.9 / 53.8 & {\color{red} -3.8} / {\color{blue} +4.5} & {\color{red} -3.7} / {\color{red} -3.1} & {\color{red} -6.9} / {\color{blue} +1.3} & {\color{red} -3.0} / {\color{red} -2.6}\\ 
	Claude 3 Haiku & 83.6 / 73.5 & {\color{red} -7.3} / {\color{red} -9.6} & {\color{red} -5.0} / {\color{red} -5.9} & {\color{red} -18.0} / {\color{red} -15.4} & {\color{red} -2.4} / {\color{blue} +6.5}\\ 
	WS-BERT & 80.9 / 80.3 & {\color{blue} +1.0} / {\color{blue} +1.1} & {\color{blue} +0.1} / {\color{blue} +0.1} & {\color{blue} +1.0} / {\color{blue} +1.1} & {\color{blue} +1.0} / {\color{blue} +1.1}\\ 
	\hline 
	\multicolumn{6}{c}{SemEval 2016 Task 6 (Accuracy~/~Macro F1 in \%)}\\ 
 	\hline 
	Llama-3.2-3B & 63.4 / 46.3 & {\color{red} -2.4} / {\color{red} -1.1} & {\color{red} -4.6} / {\color{red} -8.1} & {\color{red} -16.3} / {\color{red} -27.9} & {\color{red} -22.9} / {\color{red} -18.0}\\ 
	Llama-3.1-8B & 68.9 / 63.3 & {\color{red} -5.7} / {\color{red} -5.3} & {\color{red} -6.3} / {\color{red} -7.0} & {\color{red} -12.6} / {\color{red} -16.8} & {\color{red} -9.5} / {\color{red} -8.4}\\ 
	Qwen2.5-3B & 45.7 / 43.3 & {\color{red} -2.6} / {\color{red} -7.1} & {\color{blue} +3.8} / {\color{red} -0.3} & {\color{blue} +4.6} / {\color{red} -0.8} & {\color{blue} +0.1} / {\color{red} -4.5}\\ 
	Qwen2.5-7B & 58.7 / 53.2 & {\color{red} -0.3} / {\color{red} -3.5} & {\color{red} -1.0} / {\color{red} -5.1} & {\color{red} -4.4} / {\color{red} -11.2} & {\color{red} -1.6} / {\color{red} -8.4}\\ 
	Qwen2.5-14B & 71.7 / 67.1 & {\color{red} -2.2} / {\color{red} -2.8} & {\color{red} -1.6} / {\color{red} -3.9} & {\color{red} -4.2} / {\color{red} -9.1} & {\color{red} -3.4} / {\color{red} -2.8}\\ 
	Qwen2.5-32B & 70.1 / 67.2 & {\color{red} -0.9} / {\color{red} -1.8} & {\color{red} -1.0} / {\color{red} -2.1} & {\color{red} -2.9} / {\color{red} -6.2} & {\color{red} -0.7} / {\color{red} -1.3}\\ 
	GPT-4o mini & 74.1 / 67.6 & {\color{red} -2.5} / {\color{red} -5.4} & {\color{red} -3.4} / {\color{red} -6.4} & {\color{red} -4.6} / {\color{red} -9.6} & {\color{red} -2.5} / {\color{red} -6.3}\\ 
	Claude 3 Haiku & 71.8 / 67.6 & {\color{red} -1.7} / {\color{red} -5.1} & {\color{blue} +2.8} / {\color{red} -4.5} & {\color{red} -8.6} / {\color{red} -20.5} & {\color{blue} +0.4} / {\color{red} -4.7}\\ 
	WS-BERT & 70.9 / 57.2 & {\color{red} -1.2} / {\color{red} -2.3} & {\color{red} -0.1} / {\color{red} -0.7} & {\color{red} -1.4} / {\color{red} -2.5} & {\color{red} -1.4} / {\color{red} -2.4}\\ 
	\hline 
	
    \end{tabular}
    \caption{Use of external information degrades LLM stance detection performance. In each row, {\it No Info} stands for the performance when there is no external information; other columns contain the performance relative to {\it No Info}, in the presence of external information: Wikipedia, web search, Wikipedia with synthetic Against bias, and Wikipedia with synthetic Favor bias, respectively. We color positive and negative changes in {\color{blue} blue} and {\color{red} red}, respectively.}
    \label{tab:accuracy-f1-change}
\end{table*}    

\subsection{Research Questions}

\noindent \textbf{RQ1. Effects of external information on performance.} We first ask \textit{how the stance detection performance of LLMs changes as external information is introduced}. We evaluate LLMs when external information (from either Wikipedia or web search) is given, relative to when no external information is available. All eight LLMs are evaluated without further training, while WS-BERT is trained using the configuration in \citet{infusing}.

\noindent \textbf{RQ2. Analysis on information characteristics.} To explain observations in RQ1, we study \textit{how external information stance and sentiment are related to a model's predictions} and \textit{whether external information length correlates with performance changes}.
Hypothesizing that the {\it perceived} stance or sentiment (positive/negative/none) by a model is what influences the model's final prediction, we consider such perception instead of information stance or sentiment {\it in isolation} (e.g., by having a fixed reference sentiment analysis model). We compare the proportion of Tweets for which the model's prediction is aligned with the external information stance or sentiment, with and without such information\footnote{For sentiments, this means information with a {\it positive} sentiment may lead the stance prediction towards {\it favorable}.}. This is captured through the net adoption rate ($\NAR$) in Equation~\ref{eq:tendency}. We also compare the number of changes leading to incorrect (harmful) and correct predictions through the harmful adoption rate ($\HAR$).
Given a model $M$ for a target $T$ and external information $t$, we have
\begin{equation}
    \begin{aligned}
    \NAR (t, T) &= P(M(x|t, T)=s)\\
    &\quad -P(M(x|\varnothing, T)=s)\\ 
    \HAR (t, T) &= P(M(x|t, T)=s,g(x)\neq s,\\&\qquad\qquad  M(x|\varnothing, T)\neq s)\\&\quad -P(M(x|t, T)=s,g(x)= s,\\&\qquad\qquad  M(x|\varnothing, T)\neq s)\, ,
    \end{aligned}
    \label{eq:tendency}
\end{equation}
where $M(x|t,T)$ is the stance detected by $M$ for a text $x$ given information $t$ and target $T$; $s=M(t|\varnothing, T)$ is the stance or sentiment (depending on the context) $M$ detects for $t$; $P(c)$ is the proportion of Tweets associated with $T$ satisfying a condition $c$; $g(x)$ is the ground truth stance of $x$. When $\NAR>0$, the model {\it adopts} the stance or sentiment of the information. When $\HAR>0$, such adoptions are more {\it harmful} than helpful.
As for length, the number of tokens in each piece of information is counted using tiktoken~\citep{tiktoken}. 
We analyze one model from each model family (Llama-3.1-8B, Qwen2.5-7B, GPT-4o mini, and Claude 3 Haiku) for this experiment.

\begin{figure*}[!h]
    \centering
    \includegraphics[width=.49\linewidth]{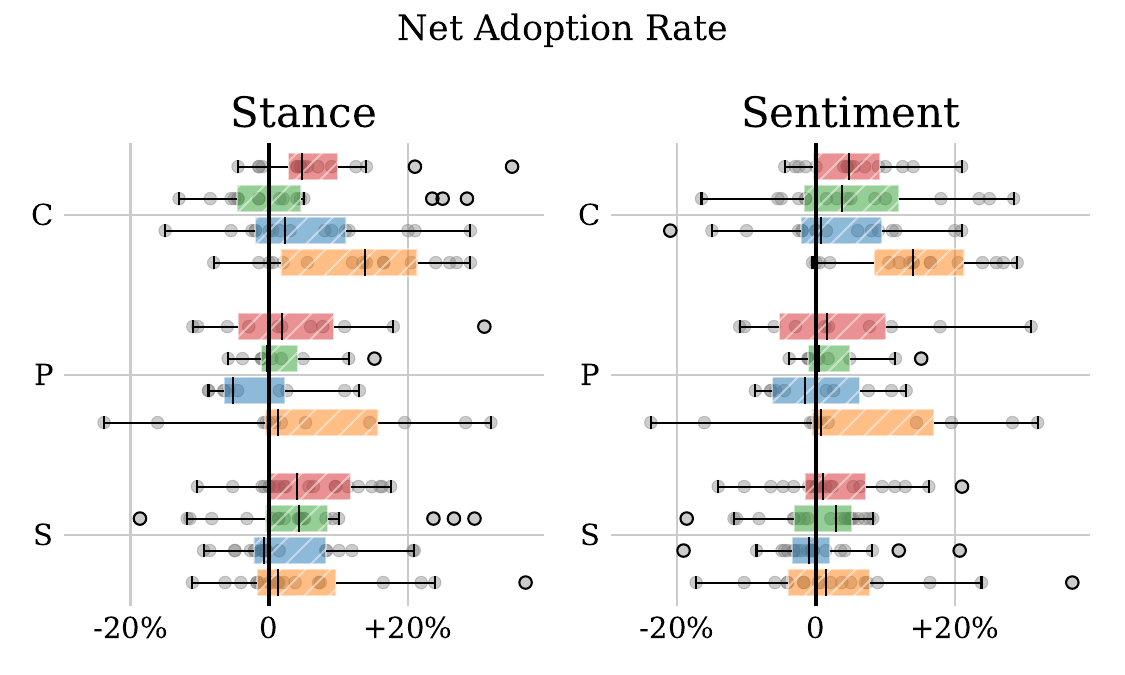}
    \includegraphics[width=.49\linewidth]{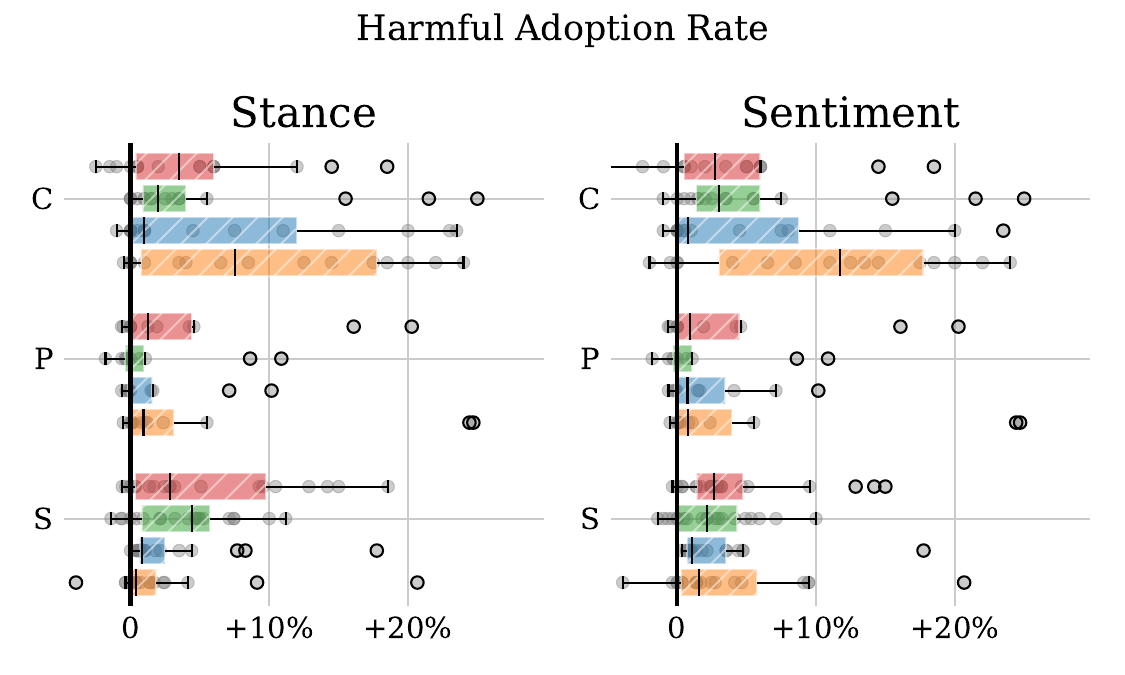}
    \includegraphics[width=.53\linewidth]{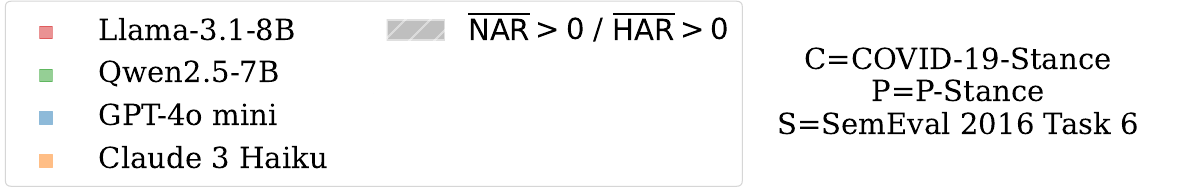}
    \caption{{\it Left}: Predictions are more likely to move towards information stance and sentiment than they are to move away. {\it Right}: When a model adopts information stance or sentiment, predictions are more likely to be wrong (harmful) than to be correct (helpful).}
    \label{fig:tendency}
\end{figure*}

\noindent \textbf{RQ3. Variations in prompting strategies.} As LLMs are known to be sensitive to prompting variations~\cite{promptdesign}, we identify \textit{how performance changes with zero-shot~\citep{cot} and few-shot~\citep{few-shot-cot} chain-of-thought (CoT) prompting}. Our prompts explicitly instruct the model to incorporate the provided external information without adopting its stance. Experiments are conducted using Llama-3.1-8B, Qwen2.5-7B, GPT-4o mini, and Claude 3 Haiku.

\noindent \textbf{RQ4. Fine-tuning.} Finally, we examine \textit{how fine-tuning affects our findings}, as performance changes may vary when models are fine-tuned alongside with external information. For each target and information type, we train two low-rank adapters (LoRA)~\citep{lora} of ranks 16 and 32. We fine-tune both Llama-3.1–8B and Qwen2.5–7B for 3 epochs with a batch size of 8 and a learning rate of $1\times10^{-4}$. This results in 96 LoRAs in total. To ensure fair comparisons with identical training settings, closed-source models are not included.

\section{Results and Discussion}
\label{section:results}

\subsection{Effects of External Information on Performance}

Table~\ref{tab:accuracy-f1-change} shows the performance of all models with different types of external information, relative to their performance without it.\footnote{Note that macro F1 is an unweighted average across tasks in each dataset while accuracy is weighted.} While there are variations depending on the model, information type, and dataset, we see an overall trend of performance degradation. Using information from Wikipedia, the most extreme drop in macro F1 is observed for Llama-3.2-3B for P-Stance, by {\color{red}15.7\%}, which becomes more severe when synthetic biases are introduced (by up to {\color{red} 27.9\%} for Llama-3.2-3B, SemEval 2016 Task 6, and Against bias); with web-search information, the steepest drop in macro F1 reaches {\color{red}18.3\%} for Claude 3 Haiku on COVID-19-Stance.
This behavior of LLMs contrasts with the fine-tuned WS-BERT, for which the performance generally increases, consistent with \citet{infusing}.\footnote{Note that WS-BERT's performance decreases, though by a small margin, on the SemEval 2016 Task 6 dataset.} Together, these results suggest that \textit{external information often decreases LLMs' stance detection performance}.

\subsection{Analysis on Information Characteristics}
\label{sec:results:analysis-information}

To gain insights into why external information often reduces performance, we consider the effects of information stance, sentiment, and length. Figure~\ref{fig:tendency} visualizes $\NAR$ and $\HAR$ as defined in Equation~\ref{eq:tendency}. We observe that {\it in the presence of external information, a model is likely to adopt the stance of such information in its predictions}: the mean net adoption rate is positive in {\color{blue}11 out of 12} model–dataset combinations for stance, and in {\color{blue}10 out of 12} combinations for sentiment.
Figure~\ref{fig:tendency} ({\it Right}) shows that {\it when such adoption happens, more incorrect predictions are made than correct ones}: the mean harmful adoption rates are positive for all models and datasets.

\begin{figure}[!h]
    \centering
    \includegraphics[width=\linewidth]{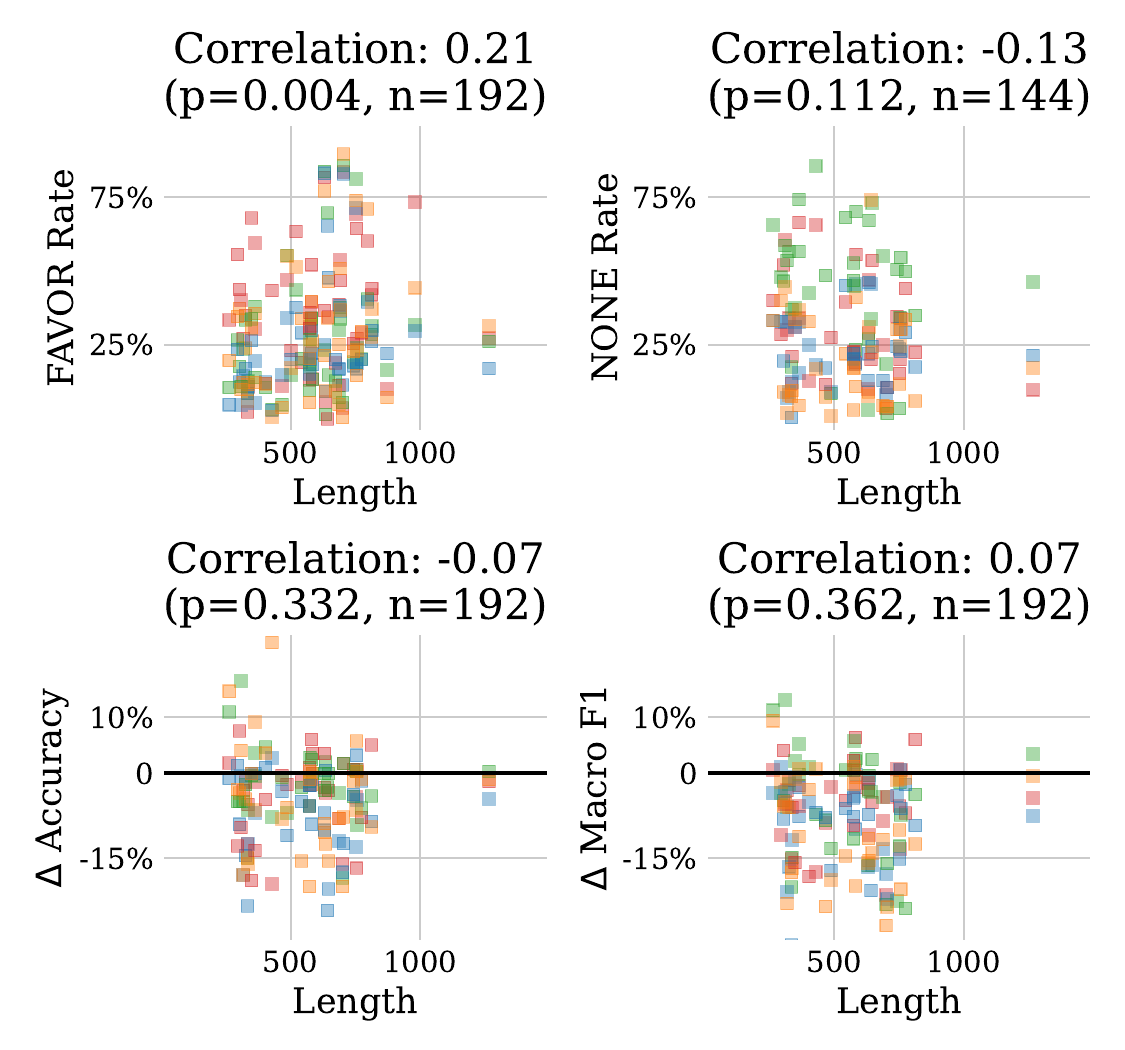}
    \includegraphics[width=\linewidth]{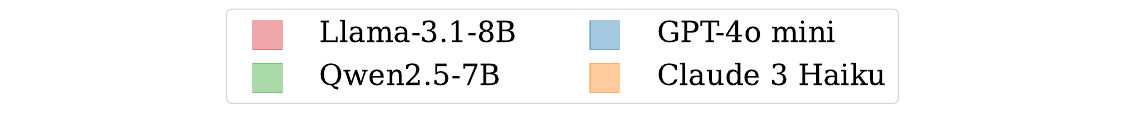}
    \caption{{\it Top}: There is some correlation between external information length and FAVOR predictions but not NONE predictions.
    {\it Bottom}: There is no correlation between external information length and performance change due to such information. Details on correlation calculations are in Appendix~\ref{app:length-correlations}.}
    \label{fig:length-correlation}
\end{figure}

In Figure~\ref{fig:length-correlation}, we visualize the correlation between information length and rates of predicted classes as well as relative performance. We observe a {\it correlation between length and the rate of FAVOR predictions but no significant correlation in other cases}. These results align with the work of \citet{promptdesign}, where the correlation between prompt length and task performance is minimal across many multiple-choice and classification tasks.

\subsection{Variations in Prompting Strategies}
\label{sec:results:prompt-strategy}

\begin{figure}[!h]
    \centering
    \includegraphics[width=\linewidth]{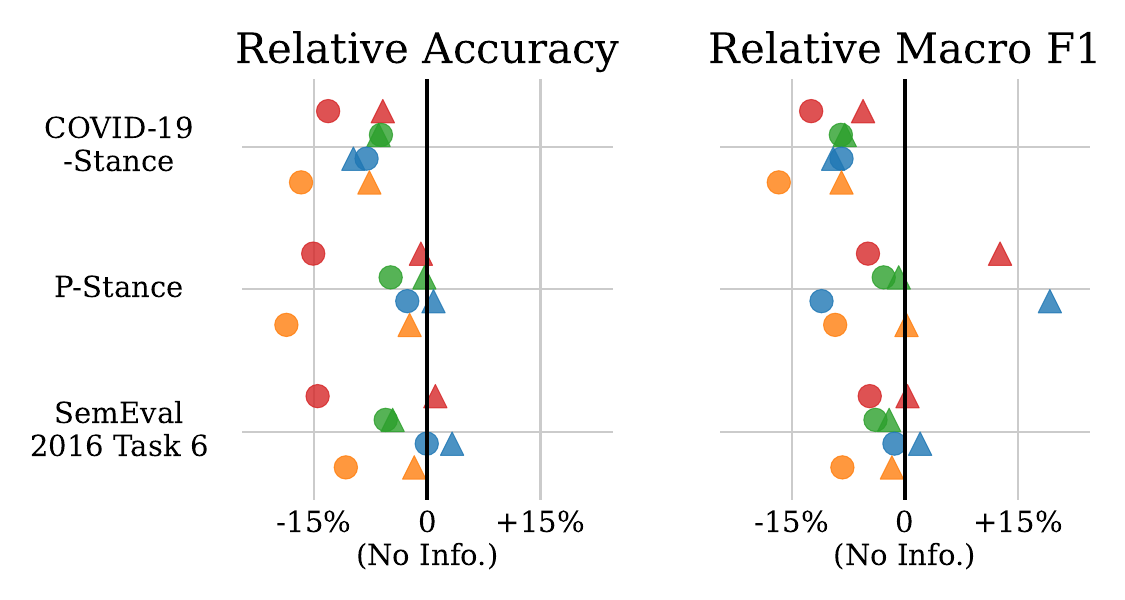}
    \includegraphics[width=\linewidth]{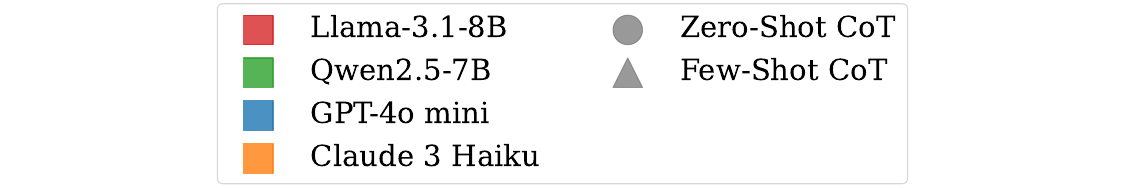}
    \caption{Use of external information degrades stance detection performance when zero-shot and few-shot CoTs are employed. We plot the average results for all types of information.}
    \label{fig:prompt-strategy}
\end{figure}

Figure~\ref{fig:prompt-strategy} illustrates the performance changes of Llama-3.1-8B, Qwen2.5-7B, GPT-4o-mini, and Claude 3 Haiku with zero-shot and few-shot CoT reasoning. Surprisingly, in most cases, {\it external information also decreases the performance of the considered LLMs with CoTs}. The steepest average F1 drops are by {\color{red}16.7\%} (with Claude 3 Haiku and COVID-19-Stance) for zero-shot CoT and {\color{red}9.5\%} (with GPT-4o-mini and COVID-19-Stance) for few-shot CoT.

\subsection{Fine-Tuning}
\label{sec:results:fine-tuning}

\begin{figure}[!h]
    \centering
    \includegraphics[width=\linewidth]{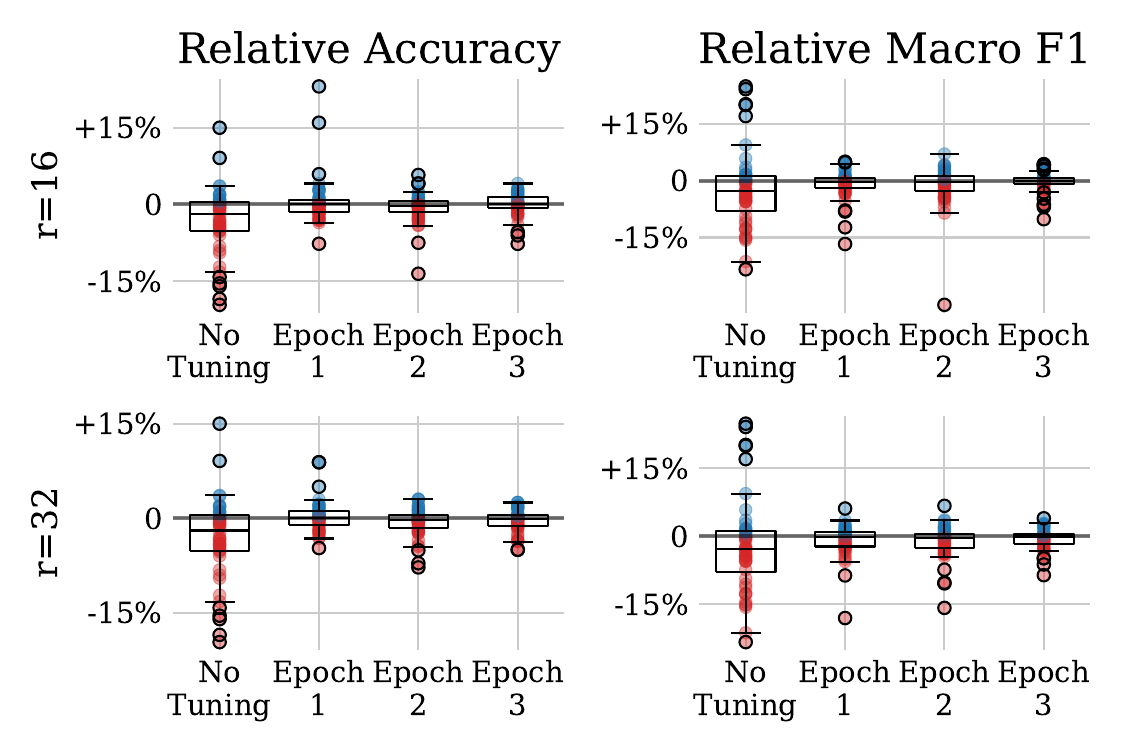}
    \caption{Models become more robust but do not completely eliminate the negative effects of external information through fine-tuning. Performance visualized is relative to the no-information setting. {\it Top}: Results for LoRA rank 16. {\it Bottom}: Results for LoRA rank 32.}
    \label{fig:fine-tuning}
\end{figure}

To examine how fine-tuning shapes the role of external information, we visualize the relative performance of models through 3 epochs of fine-tuning in Figure~\ref{fig:fine-tuning}. Overall, we observe a monotonic increase with variances contracting with more epochs.
Nevertheless, even at the third epoch, we still observe {\color{red} 25 out of 48} rank-32 instances falling below zero for both relative accuracy and relative macro F1 (see Appendix~\ref{app:fine-tuning} for more details). This means that fine-tuned LLMs still \textit{do not benefit from external information} in many cases.

\section{Conclusion}

We investigated the question of whether external information can benefit LLMs for stance detection. Contradicting previous literature on BERT-based stance detection with external information, our experiments indicated that such information can actually harm the performance of LLMs. We also verified that this phenomenon is partly caused by LLMs being biased by the stance or sentiment they perceive in external information. Furthermore, CoT prompting is of little benefit, while fine-tuning lessens but does not completely alleviate this problem. Given such observations, we call for more consideration of bias factors in LLM stance detection and natural language reasoning at large.

\section{Limitations}

Our work provided a systematic evaluation of how external information can affect the performance of LLM stance detection systems. This research can serve as a foundation for a number of crucial future research directions.

Our analysis involving information characteristics represents one of many possible perspectives and levels of depth for interpreting the main results. For instance, another perspective could involve probing the attention heads of models~\citep{marksgeometry, linear-politics}.  We hope future work will explore such complementary approaches in depth.

Additionally, even though we have conducted experiments on prompting variations and fine-tuning, there could still be advanced methods which may more efficiently improve the stance detection performance of LLMs under external information. Since our focus is on the analysis of effects external information can have on LLMs, investigations on more optimal mitigation measures are left open for future research.

\section{Ethical Considerations}

Given the tendency of LLMs to be biased by the stance of external information, as investigated in our paper, it is possible for malicious actors to manipulate open information sources such as Wikipedia to alter the outputs of LLM stance detection systems. We caution against the use of external information without proper curation of the information source and also encourage future research on mitigation measures.

Furthermore, even though Tweets in the datasets we utilized have been anonymized by their respective authors~\citep{covid19stance,pstance,semeval}, their content might contain offensive language against targets. Our work reports aggregated statistics and analysis from such data but does not present any offensive information individually.

\section*{Acknowledgements}

This work was supported by the National Research Foundation of Korea (RS-2022-NR068758) and Underscore.
\bibliography{2025-acl-camera-ready}

\appendix

\renewcommand{\thetable}{A\arabic{table}}
\renewcommand{\thefigure}{A\arabic{figure}}
\setcounter{table}{0}
\setcounter{figure}{0}

\section{Details of Data and Models}
\label{app:llms}

All of the datasets we use are made publicly available by their respective authors~\citep{covid19stance, pstance, semeval}. Readers can refer to the orignal papers for instructions on how to access the data. Their statistics are included in Table~\ref{tab:data-statistics}. Note that since the SemEval data only has training and testing sets, we perform stratified sampling with the \texttt{scikit-learn} library to use 20\% of the training sets as validation sets.
\begin{table*}[!h]
    \centering
    \begin{tabular}{c|ccc|ccc|ccc}
    \hline
    Target & \multicolumn{3}{|c|}{Train} & \multicolumn{3}{|c|}{Val} & \multicolumn{3}{|c}{Test} \\
    & Favor & Against & None & Favor & Against & None & Favor & Against & None\\
    \hline
    \multicolumn{10}{c}{COVID-19-Stance} \\
    \hline 
    Face Masks & 531 & 512 & 264 & 81 & 78 & 41 & 81 & 78 & 41 \\
    Fauci & 388 & 480 & 596 & 52 & 65 & 83 & 52 & 65 & 83 \\
    School Closures & 409 & 166 & 215 & 103 & 42 & 55 & 103 & 42 & 55 \\
    Stay at Home Orders & 136 & 284 & 552 & 27 & 58 & 115 & 27 & 58 & 115 \\
    \hline
    \multicolumn{10}{c}{P-Stance} \\
    \hline 
    Bernie Sanders & 2858 & 2198 & 0 & 350 & 284 & 0 & 343 & 292 & 0 \\
    Joe Biden & 2552 & 3254 & 0 & 328 & 417 & 0 & 337 & 408 & 0 \\
    Donald Trump & 2937 & 3425 & 0 & 374 & 440 & 0 & 352 & 425 & 0 \\
    \hline
    \multicolumn{10}{c}{SemEval 2016 Task 6} \\
    \hline 
    Atheism & 74 & 243 & 93 & 18 & 61 & 24 & 32 & 160 & 28 \\
    Climate Change & 170 & 12 & 134 & 42 & 3 & 34 & 123 & 11 & 35 \\
    Feminist Movement & 168 & 262 & 101 & 42 & 66 & 25 & 58 & 183 & 44 \\
    Hillary Clinton & 89 & 289 & 133 & 23 & 72 & 33 & 45 & 172 & 78 \\
    Abortion & 84 & 267 & 131 & 21 & 67 & 33 & 46 & 189 & 45 \\
    \hline
    \end{tabular}
    \caption{The number of samples in each target and split of the datasets. ``Climate Change" is short for ``Climate Change is a Real Concern". ``Abortion" is short for ``Legalization of Abortion".}
    \label{tab:data-statistics}
\end{table*}

The LLMs we utilize include Claude 3 Haiku (snapshot {\tt 20240307})~\citep{anthropic_claude_2024},
GPT-4o mini (version {\tt 2024-07-18})~\citep{gpt-4o},
Llama-3.2-3B-Instruct and Llama-3.1-8B-Instruct~\citep{llama}, and
Qwen2.5-\{3B, 7B, 14B, 32B\}-Instruct~\citep{qwen}. We used 4-bit quantizations for all Llama and Qwen models provided by Unsloth~\citep{unsloth}. We also made use of Unsloth's fine-tuning library. Inference is done through the vLLM library~\citep{vllm}.

All of our inference temperatures are set to zero, yielding negligible performance stochasticity, because of which results are reported with single runs. Our hardware was 4$\times$ NVIDIA RTX A5000, with which a single run over all of our training and evaluation (for both LLMs and BERT models) takes approximately 25 GPU hours. Our evaluation through OpenAI API (for GPT-4o mini)
and Anthropic API (for Claude 3 Haiku) cost approximately 75 USD.

\section{Prompts}
\label{app:prompts}

\noindent {\bf Non-CoT Stance Detection Prompt.} For non-CoT stance detection, all models are prompted with the following instruction:

\begin{tcolorbox}[colback=cyan!5, arc=0mm, boxrule=0mm, colframe=cyan!5, left=0pt, right=0pt]
\noindent\texttt{USER: You are given the following text: \{text\}. What is the stance of the text towards the target `\{target\}'? {\color{red} The following information can be helpful: \{wiki\}}. Options: \{options\}. Do not explain. Just provide the stance in a single word.\\
ASSISTANT:}
\end{tcolorbox}

Here {\tt\{wiki\}} stands for the external information excerpt that can also be empty, in which case the sentence in {\color{red}red} is omitted. Meanwhile, {\tt \{options\}} is \texttt{[FAVOR, AGAINST]} for P-Stance and \texttt{[FAVOR, AGAINST, NONE]} for COVID-19-Stance and SemEval 2016 Task 6. USER and ASSISTANT are replaced with tokens corresponding to the chat prompt template of each model.

\noindent {\bf Zero-Shot CoT Stance Detection Prompt.} Following \citet{cot}, we prompt models as follows:

\begin{tcolorbox}[colback=cyan!5, arc=0mm, boxrule=0mm, colframe=cyan!5, left=0pt, right=0pt, breakable]
\noindent\texttt{USER: You are given the following text: \{text\}. What is the stance of the text towards the target `\{target\}'? {\color{red} Integrate \color{red}  the following external information and do NOT automatically adopt the stance of it: \{wiki\}}. Options: \{options\}\\ 
ASSISTANT: Let's think step by step.}
\end{tcolorbox}

Again, the sentence in {\color{red}red} is omitted when there is no external information.

\noindent {\bf Few-Shot CoT Stance Detection Prompt.} Our prompt for the few-shot CoT setting \citep{few-shot-cot} has the following format: 

\begin{tcolorbox}[colback=cyan!5, arc=0mm, boxrule=0mm, colframe=cyan!5, left=0pt, right=0pt, breakable]
\noindent\texttt{USER: You are given the following text: \{FAVOR text\}. What is the stance of the text towards the target `\{target\}'? {\color{red} Integrate the following external information and do NOT automatically adopt the stance of it: \{wiki\}}. Options: \{options\}\\
ASSISTANT: \{FAVOR CoT\}\\~\\
USER: You are given the following text: \{AGAINST text\}. What is the stance of the text towards the target `\{target\}'? {\color{red} Integrate the following external information and do NOT automatically adopt the stance of it: \{wiki\}}. Options: \{options\}\\
ASSISTANT: \{AGAINST CoT\}\\~\\
USER: You are given the following text: \{NONE text\}. What is the stance of the text towards the target `\{target\}'? {\color{red} Integrate the following external information and do NOT automatically adopt the stance of it: \{wiki\}}. Options: \{options\}\\
ASSISTANT: \{NONE CoT\}\\~\\
USER: You are given the following text: \{text\}. What is the stance of the text towards the target `\{target\}'? {\color{red} Integrate the following external information and do NOT automatically adopt the stance of it: \{wiki\}}. Options: \{options\}\\
ASSISTANT:
}
\end{tcolorbox}

The sentences in {\color{red}red} are omitted when there is no external information. Here {\tt \{FAVOR text\}}, {\tt \{AGAINST text\}}, and {\tt \{NONE text\}} are Tweets with FAVOR, AGAINST, and NONE ground truth labels, respectively. Meanwhile, {\tt\{FAVOR CoT\}}, {\tt\{AGAINST CoT\}}, and {\tt  \{NONE CoT\}} are our CoT examples for each class. The last in-context example, for the NONE class, is not included for the P-Stance dataset, which has no such class.

The authors wrote the CoT examples themselves for each target and information type (Wikipedia, Wikipedia with Against bias, Wikipedia with Favor bias, web search, or no information).

\noindent {\bf Web Search Prompt.} For the COVID-19-Stance dataset, we search using the prompt

\begin{tcolorbox}[colback=cyan!5, arc=0mm, boxrule=0mm, colframe=cyan!5, left=0pt, right=0pt, breakable]
\noindent \texttt{
Provide a summary of \{target\} in the context of COVID-19.
} 
\end{tcolorbox}

For the P-Stance dataset, the prompt is

\begin{tcolorbox}[colback=cyan!5, arc=0mm, boxrule=0mm, colframe=cyan!5, left=0pt, right=0pt, breakable]
\noindent \texttt{
Provide a summary of \{target\} in the context of the 2020 U.S. presidential election.
}
\end{tcolorbox}

For the SemEval 2016 Task 6 dataset, we search using the prompt

\begin{tcolorbox}[colback=cyan!5, arc=0mm, boxrule=0mm, colframe=cyan!5, left=0pt, right=0pt, breakable]
\noindent \texttt{
Provide a summary of \{target\}.
}
\end{tcolorbox}

The prompts for COVID-19-Stance and P-Stance are written while taking into account the context of each dataset (COVID-19 for the former and the 2020 U.S. presidential election for the latter.)

\begin{table*}[!h]
    \centering
    \begin{tabular}{ccccc}
	\hline 
 	 Metric & No Tuning & Ep. 1 & Ep. 2 & Ep. 3 \\ 
	\hline 
	\multicolumn {5}{c}{LLMs, LoRA rank 16 (out of 48 instances)} \\ 
	\hline 
 	Accuracy & 32& 21& 25& 20\\ 
 	Macro F1 & 31& 27& 27& 25\\ 
 	\hline 
	\multicolumn {5}{c}{LLMs, LoRA rank 32 (out of 48 instances)} \\ 
	\hline 
 	Accuracy & 32& 17& 27& 25\\ 
 	Macro F1 & 31& 26& 28& 25\\ 
 	\hline 
	\multicolumn {5}{c}{WS-BERT (out of 24 instances)} \\ 
	\hline 
 	Accuracy & 3& 12& 4& 8\\ 
 	Macro F1 & 5& 17& 4& 10\\ 
 	\hline
    \end{tabular}
    \caption{The number of combinations of target-model-type of external information in each fine-tuning epoch for which the performance is lower than that of the same target and model without external information.}
    \label{tab:negative_instances}
\end{table*}

\begin{table*}[!ht]
    \centering
    \begin{tabular}{ccccc}
	\hline 
 	 Metric & No Tuning & Ep. 1 & Ep. 2 & Ep. 3 \\ 
	\hline 
	\multicolumn {5}{c}{LLMs, LoRA rank 16 (in \% performance)} \\ 
	\hline 
 	Accuracy Mean & -3.28 & 0.61 & -0.59 & -0.07 \\ 
 	Accuracy Std & 6.59& 4.58& 2.93& 2.30\\ 
 	Macro F1 Mean & -2.21 & -1.02 & -1.30 & -0.43 \\ 
 	Macro F1 Std & 10.46& 3.90& 5.49& 2.84\\ 
 	\hline 
	\multicolumn {5}{c}{LLMs, LoRA rank 32 (in \% performance)} \\ 
	\hline 
 	Accuracy Mean & -3.28 & 0.34 & -0.62 & -0.41 \\ 
 	Accuracy Std & 6.59& 2.47& 2.29& 1.86\\ 
 	Macro F1 Mean & -2.21 & -0.99 & -1.25 & -0.59 \\ 
 	Macro F1 Std & 10.46& 3.53& 3.75& 2.36\\ 
 	\hline 
	\multicolumn {5}{c}{WS-BERT (in \% performance)} \\ 
	\hline 
 	Accuracy Mean & -0.02 & 2.17 & 1.75 & 0.20 \\ 
 	Accuracy Std & 1.46& 15.08& 2.48& 1.72\\ 
 	Macro F1 Mean & -0.15 & -1.29 & 3.56 & 0.13 \\ 
 	Macro F1 Std & 1.32& 5.95& 5.57& 2.47\\ 
 	\hline
    \end{tabular}
    \caption{The mean and standard deviation of relative performance in each fine-tuning epoch.}
    \label{tab:mean-and-std}
\end{table*}

\noindent {\bf Synthetic Stance Injection Prompt.}
We inject synthetic Favor and Against stances into Wikipedia information excerpts using the following prompt:

\begin{tcolorbox}[colback=cyan!5, arc=0mm, boxrule=0mm, colframe=cyan!5, left=0pt, right=0pt, breakable]
\noindent\texttt{USER: You are given the following Wikipedia entry: \{wiki\}. Rewrite the Wikipedia entry to have the stance `\{stance\}' towards the target '\{target\}'. Be discreet and do not change the factual content.\\
ASSISTANT:}
\end{tcolorbox}

\noindent {\bf Perceived Stance and Sentiment Prompt.}
We collect LLMs' perceived sentiment of external information using the prompt
 
\begin{tcolorbox}[colback=cyan!5, arc=0mm, boxrule=0mm, colframe=cyan!5, left=0pt, right=0pt, breakable]
\noindent\texttt{USER: You are given the following text: \{information\}. What is the sentiment of the text? Options: \{options\}. Do not explain. Just provide the sentiment in a single word.\\
ASSISTANT:}
\end{tcolorbox}

{\{options\}} is {\tt [POSITIVE, NEGATIVE, NONE]} for COVID-19-Stance and SemEval 2016 Task 6 and {\tt [POSITIVE, NEGATIVE]} for P-Stance. 

Similarly, the perceived stance of information towards a target is collected using the prompt

\begin{tcolorbox}[colback=cyan!5, arc=0mm, boxrule=0mm, colframe=cyan!5, left=0pt, right=0pt, breakable]
\noindent\texttt{USER: You are given the following text: \{information\}. What is the stance of the text towards the target `\{target\}'? Options: \{options\}. Do not explain. Just provide the stance in a single word.\\
ASSISTANT:}
\end{tcolorbox}

{\{options\}} is {\tt [FAVOR, AGAINST, NONE]} for COVID-19-Stance and SemEval 2016 Task 6 and {\tt [FAVOR, AGAINST]} for P-Stance.

\section{LLM Output Validation}
\label{app:validation}

For the non-CoT setting, if the output is within \{\texttt{favor, favour, favorable, favourable}\} (non-case-sensitive), we register the final answer as `FAVOR'. If the output is \texttt{against}, we register `AGAINST'. For outputs that are within \{\texttt{none, neutral}\}, we register `NONE'. Other answers are considered invalid. In the case of stance detection with 2-classes, applying to P-Stance among our datasets, \texttt{none} and \texttt{neutral} outputs are invalid.

For CoT settings, we follow suggestions by \citet{cot} and conduct a second round of prompting for answer extraction. The answer extraction prompt is a concatenation of the original prompt, the generated CoT, and a trigger sentence. We use "{\tt Therefore, among FAVOR, AGAINST, and NONE, the final answer is }" for COVID-19-Stance and SemEval 2016 Task 6 and "{\tt Therefore, between FAVOR and AGAINST, the final answer is }" for P-Stance as trigger sentences. Again, following \citep{cot}, we parse the final answer by getting the first match after "the final answer is" that is within the set \{{\tt favor}, {\tt favour}, {\tt against}, {\tt none}, {\tt neutral}\} (non-case-sensitive). For P-Stance, {\tt none} and {\tt neutral} are considered invalid answers.

Sentiment outputs are similarly validated, with {\tt favor} (and variations) and {\tt against} replaced with {\tt positive} and {\tt negative}, respectively.

\section{Notes on Length Correlations}
\label{app:length-correlations}

In Figure~\ref{fig:length-correlation}, note that FAVOR prediction rates are calculated after excluding NONE predictions for a fair comparison among the three datasets. For NONE prediction rates, we exclude P-Stance, which has no such label. There is no AGAINST prediction rate visualization since its correlation with information length is precisely the additive inverse of the FAVOR counterpart: $\rho (1-X, Y)=-\rho(X,Y)$, where $X$ is the FAVOR prediction rate, $1-X$ is the AGAINST prediction rate, and $Y$ denotes information length.

\section{Performance Changes with Fine-tuning}
\label{app:fine-tuning}

Tables~\ref{tab:negative_instances} and \ref{tab:mean-and-std} show the number of instances with negative relative performance as well as relative performance in each fine-tuning epoch.

\end{document}